\documentclass[10pt,twocolumn,letterpaper]{article}

\usepackage{cvpr}
\usepackage{times}
\usepackage{epsfig}
\usepackage{graphicx}
\usepackage{amsmath,mathtools}
\usepackage{amssymb}
\usepackage{subcaption}

\usepackage{algorithm}
\usepackage{algpseudocode}
\algrenewcommand\algorithmicindent{.9em}%

\usepackage[pagebackref=true,breaklinks=true,letterpaper=true,colorlinks,bookmarks=false]{hyperref}

\cvprfinalcopy 

\def\cvprPaperID{****} 

\ifcvprfinal\fi

\newcommand{\nfeat}{\mathcal{N}_\textit{feat}}
\newcommand{\ntask}{\mathcal{N}_\textit{task}}
\newcommand{\nq}{\mathcal{N}_\textit{q}}
\newcommand{\vframe}[2]{\mathbf{I}^{#1}_{#2}}
\newcommand{\vseg}[1]{\mathbf{S}_{#1}}
\newcommand{\feat}[2]{\mathbf{f}^{#1}_{#2}}

\newcommand{\kfint}{l}
\newcommand{\agg}{\mathcal{A}}
\newcommand{\ff}[1]{\mathbf{M}_{#1}}

\newcommand{\tnew}{\text{new}}
\newcommand{\told}{\text{old}}
\newcommand{\tcur}{\text{cur}}
\newcommand{\timp}{\text{imp}}
\newcommand{\ttsk}{\text{task}}
\newcommand{\tkey}{\text{key}}
\newcommand{\tref}{\text{ref}}

\newcommand{\mmgt}{g}
\newcommand{\warp}{\mathcal{W}}
\newcommand{\warpp}[2]{\warp(#1, #2)}
\newcommand{\scl}{\mathbf{S}}
\newcommand{\flow}{\mathcal{F}}
\newcommand{\flowp}[2]{\flow(#1, #2)}

\newcommand{\cmpx}[1]{O(#1)}

\begin{document}

\title{Impression Network for Video Object Detection}

\author{Congrui Hetang\thanks{This work is done when Congrui Hetang and Shaohui Liu are interns at SenseTime} \qquad Hongwei Qin \qquad Shaohui Liu$^{*}$ \qquad Junjie Yan\\
SenseTime\\
{\tt\small \{hetangcongrui, qinhongwei, liushaohui, yanjunjie\}@sensetime.com}
}

\maketitle

\begin{abstract}
Video object detection is more challenging compared to image object detection. Previous works proved that applying object detector frame by frame is not only slow but also inaccurate. Visual clues get weakened by defocus and motion blur, causing failure on corresponding frames. Multi-frame feature fusion methods proved effective in improving the accuracy, but they dramatically sacrifice the speed. Feature propagation based methods proved effective in improving the speed, but they sacrifice the accuracy. So is it possible to improve speed and performance simultaneously? 

Inspired by how human utilize impression to recognize objects from blurry frames, we propose Impression Network that embodies a natural and efficient feature aggregation mechanism. In our framework, an impression feature is established by iteratively absorbing sparsely extracted frame features. The impression feature is propagated all the way down the video, helping enhance features of low-quality frames. This impression mechanism makes it possible to perform long-range multi-frame feature fusion among sparse keyframes with minimal overhead. It significantly improves per-frame detection baseline on ImageNet VID while being 3 times faster (20 fps). We hope Impression Network can provide a new perspective on video feature enhancement. Code will be made available. 
\end{abstract}

\begin{figure*}[t]
	\begin{center}
		\includegraphics[width=1.0\linewidth]{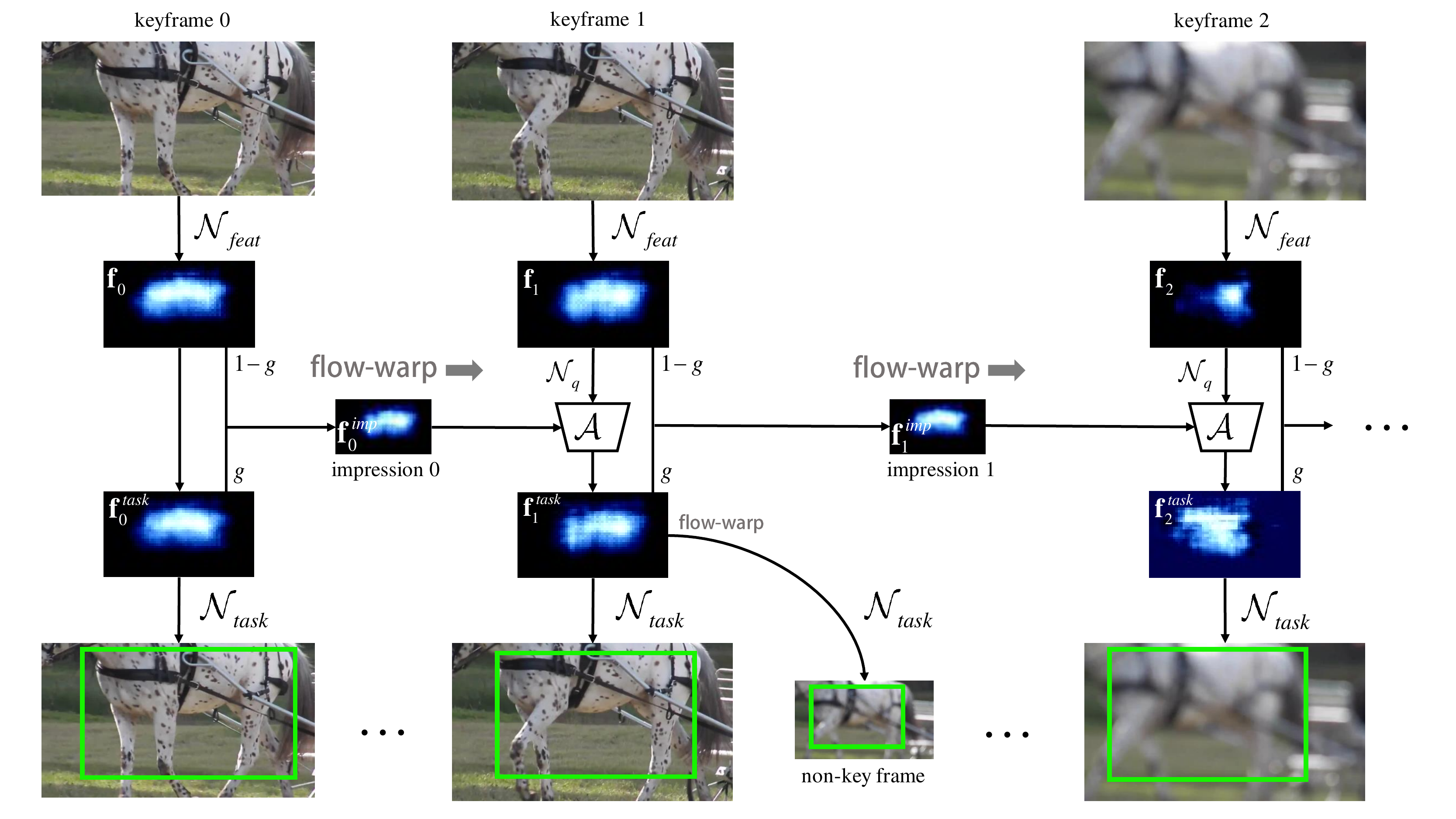}
	\end{center}
	\caption{Impression Network inference pipeline. Only keyframes of the first 3 segments are shown. The $581^{\text{st}}$ channel (sensitive to horse) of the top feature map is visualized. The detection at keyframe 2 should have failed due to defocus (Figure~\ref{fig.comp_baseline}), but the impression feature brought information from previous frames and enhanced $\feat{}{2}$, thus $\ntask$ was still able to predict correctly.}
	\label{fig.inference}
\end{figure*}

\section{Introduction}
\begin{figure}
	\begin{center}
		\includegraphics[width=1.0\linewidth]{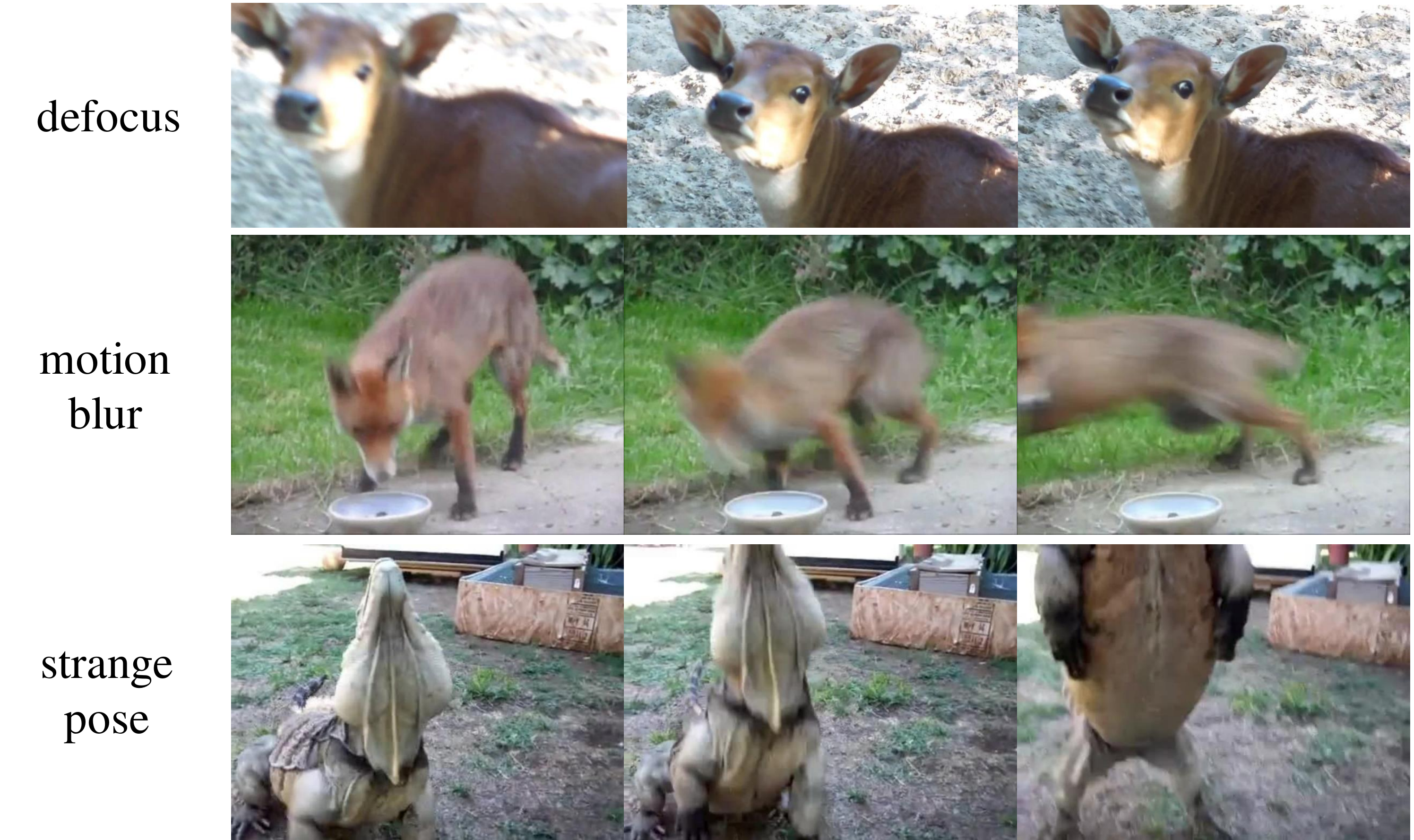}
	\end{center}
	\caption{Examples of deteriorated frames in videos. }
	\label{fig.challenge}
\end{figure}

Fast and accurate video object detection methods are highly valuable in vast number of scenarios. Single-image object detectors like Faster R-CNN~\cite{ren2015faster} and R-FCN~\cite{dai2016r} have achieved excellent accuracy on still images, so it is natural to apply them to video tasks. One intuitive way is applying them frame by frame on videos, but this is far from optimal. First, image detectors typically involve a heavy feature network like ResNet-101~\cite{he2016deep}, which runs rather slow (5fps) even on GPUs. This hampers their potential in real-time applications like autonomous driving and video surveillance. Second, single-image detectors are vulnerable to the common image degeneration problem in videos~\cite{zhu2017flow}. As shown in Figure~\ref{fig.challenge}, frames may suffer from defocus, motion blur, strange object positions and all sorts of deteriorations, leaving too weak visual clues for successful detections. The two problems make object detection in videos challenging.

Feature-level methods~\cite{gadde2017semantic,zhu2016deep,zhu2017flow,shelhamer2016clockwork} have addressed either one of the two problems. These methods treat single-image recognition pipeline as two stages: 1. the image is passed through a general feature network; 2. the result is then generated by a task-specific sub-network. When transferring image detectors to videos, feature-level methods seek ways to improve the feature stage, while the task network remains unchanged. The task-independence makes feature-level methods versatile and conceptually simple. To improve speed, feature-level methods reuse sparsely sampled deep features in the first stage~\cite{zhu2016deep,shelhamer2016clockwork}, because nearby video frames provide redundant information. This saves the expensive feature network inference and boosts speed to real-time level, but sacrifices accuracy. On the other hand, accuracy can be improved by multi-frame feature aggregation~\cite{zhu2017flow,liu2017quality}. This enables successful detection on low-quality frames, but the aggregation cost can be huge thus further slows down the framework. In this work, we combine the advantages of both tracks. We present a new feature-level framework, which runs at real-time speed and outperforms per-frame detection baseline.

Our method, called Impression Network, is inspired by the way how human understand videos. When there comes a new frame, humans do not forget previous frames. Instead, the impression is accumulated along the video, which helps us understand degenerated frames with limited visual clue. This mechanism is embodied in our method to enhance frame feature and improve accuracy. Moreover, we combine it with sparse keyframe feature extraction to obtain real-time inference speed. The pipeline of our method is shown in Figure~\ref{fig.inference}.

To address the redundancy and improve speed, we split a video into segments of equal length. For each segment, only one keyframe is selected for deep feature extraction. With flow-guided feature propagation~\cite{zhu2016deep,dosovitskiy2015flownet}, the key feature is reused by non-key frames to generate detection results. Based on this, we adopt our Impression mechanism to perform multi-frame feature fusion. When a key feature is extracted, it not only goes to task network, but is also
absorbed by a impression feature. The impression feature is then propagated down to the next keyframe. The task feature for the next keyframe is a weighted combination of its own feature and the impression feature, and the impression feature is updated by absorbing the feature of that frame. This process keeps going on along the whole video. In this framework, the impression feature accumulates high-quality video object information and is propagated all the way down, helping enhance incoming key features if the frames get deteriorated. It improves the overall quality of task features, thus increases detection accuracy.

The Impression mechanism also contributes to the speed. With the iterative aggregation policy, it minimized the cost of feature fusion. Previous work~\cite{zhu2017flow} has proved that, video frame features should be spatially aligned with flow-guided warping before aggregation, while flow computation is not negligible. Intuitive way requires one flow estimation for each frame being aggregated, while Impression Network only needs one extra flow estimation for adjacent segments, being much more efficient.

Without bells and whistles, Impression Network surpasses state-of-the-art image detectors on ImageNet VID~\cite{russakovsky2015imagenet} dataset. It's three times faster (20 fps) and significantly more accurate. We hope Impression Network can provide a new perspective on feature aggregation in video tasks. 

Code will be released to facilitate future research.

\section{Related Work}

\noindent
\textbf{Feature Reuse in Video Recognition:}
As shown by earlier analysis~\cite{wiskott2002slow,zou2012deep,jayaraman2016slow,zhang2012slow,sun2014dl}, consecutive video frames are highly similar, as well as their high-level convolutional features. This suggests that video sequences feature an inherent redundancy, which can be exploited to reduce time cost. In single image detectors~\cite{girshick2014rich,he2014spatial,ren2015faster,girshick2015fast,dai2016r}, the heavy feature network (encoder) is much more costly than the task sub-network (decoder). Hence, when transplanting image detectors to videos, speed can by greatly improved by reusing the deep features of frames. Clockwork Convnets~\cite{shelhamer2016clockwork} exploit the different evolve speed of features at different levels. By updating low and high level convolutional features at different frequency, it partially avoids redundant feature computation. It makes the network $1.3$ times faster, while sacrifices accuracy by $1\%\sim4\%$ due to the lack of end to end training. Deep Feature Flow~\cite{zhu2016deep} is another successful feature-level acceleration method. It cheaply propagates the top feature of sparse keyframe to other frames, achieving a significant speed-up ratio (from 5 fps to 20 fps). Deep Feature Flow requires motion estimation like optical flow~\cite{horn1981determining,brox2004high,weinzaepfel2013deepflow,revaud2015epicflow,dosovitskiy2015flownet,ilg2016flownet} to propagate features, where error is introduced and therefore brings a minor accuracy drop ($\sim 1\%$). Impression Network inherits the idea of Deep Feature Flow, but also utilizes temporal information to enhance the shared features. It's not only faster than per-frame baseline, but also more accurate. 

~\\
\noindent
\textbf{Exploiting Temporal Information in Video Tasks:}
Applying state-of-the-art still image detectors frame by frame on videos does not provide optimal result~\cite{zhu2017flow}. This is mainly due to the low-quality images in videos. Single image detectors are vulnerable to deteriorated images because they are restricted to the frame they are looking at, while ignoring the ample temporal information from other frames in the video. Temporal feature aggregation~\cite{kar2016adascan,li2017videolstm,sharma2015action,sun2015human,karpathy2014large,ballas2015delving,yue2015beyond} provides a way to utilize such information. Flow-Guided Feature Aggregation(FGFA)~\cite{zhu2017flow} aims at enhancing frame features by aggregating all frame features in a consecutive neighborhood. The aggregation weight is learned through end-to-end training. FGFA boosts video detection accuracy to a new level (from $74.0\%$ to $76.3\%$), yet it is three-times slower than per-frame solution ($1.3$ fps). This is caused by the aggregation cost. For each frame in the fusion range, FGFA requires one optical flow computation to spatially align it with the target frame, which costs even more time than the feature network. Additionally, since neighboring frames are highly similar, the exhaustive dense aggregation leads to extra redundancy. Impression Network fuses features in an iterative manner, where only one flow estimation is needed for every new keyframe. Moreover, the sparse feature sampling reduces the amount of replicated information. 

\section{Impression Network}

\begin{algorithm}[t]
	\caption{Inference algorithm of Impression Network for video object detection.}
	\small
	\begin{algorithmic}[1] 
		\State \textbf{input}: video frames $\{\vframe{}{}\}$, segment length $\kfint$
		\For{$k=0$ \textbf{to} $N$}
		
		\State $ \feat{}{k} = \nfeat(\vframe{\tkey}{k}) $ \Comment{extract keyframe feature}
		
		\If{$k=0$} \Comment{first keyframe}
		\State $ \feat{\timp}{k} = \feat{}{k} $ \Comment{initialize impression feature}
		\State $ \feat{\ttsk}{k} = \feat{}{k} $
		\Else
		\State $ 1-w_{k}, w_{k} = \text{softmax}({\nq({\feat{\timp}{k-1}}'), \nq(\feat{}{k})}) $ 
		\State $ \feat{\ttsk}{k} = (1-w_{k}) \cdot {\feat{\timp}{k-1}}' + w_{k} \cdot \feat{}{k} $ \Comment{adaptive weighting}
		\State $ \feat{\timp}{k} = (1-\mmgt) \cdot \feat{}{k} + \mmgt \cdot \feat{\ttsk}{k} $ \Comment{update impression feature}
		\EndIf
		
		\For{$j=0$ \textbf{to} $\kfint-1$} \Comment{feature propagation}
		\State $ \feat{k}{j} = \warpp{\feat{\ttsk}{k}}{\flowp{\vframe{k}{j}}{\vframe{\tkey}{k}}} $ \Comment{flow-guided warp}
		\State $ y^k_j = \ntask(\feat{k}{j}) $ \Comment{detection result}
		\EndFor
		
		\EndFor
		
		\State \textbf{output}: detection results $\{y\}$
	\end{algorithmic}
	\label{alg.inference}
\end{algorithm}

\subsection{Impression Network Inference}

Given a video, our task is to generate detection results for all its frames $ \vframe{}{k} $, $ i = 0, \ldots , N $. To avoid redundant feature computation, we split the frame sequence into segments of equal length $ \kfint $. In each segment $ \vseg{k} = \{ \vframe{}{kl}, \vframe{}{kl+1}, ... , \vframe{}{(k+1)l-1} \} $, only one frame $ \vframe{\text{\tkey}}{k} $ (by default we take the central frame $ \vframe{}{kl+\lfloor{l/2}\rfloor} $) is selected for feature extraction via the feature network $ \nfeat $. The key feature is propagated to remaining frames with flow-guided warping, where the flow field is computed by a light-weight flow network, following the practice of Deep Feature Flow~\cite{zhu2016deep}. Features of all frames are then fed into task network $ \ntask $ to generate detection results. 

In such framework, we use impression mechanism to exploit long-range, cross-segment temporal information. The inference phase of Impression Network is illustrated in Figure~\ref{fig.inference}. Each segment $ \vseg{k} $ generates three features: $ \feat{}{k} $ calculated by passing $ \vframe{\tkey}{k} $ through $ \nfeat $, $ \feat{\ttsk}{k} $ shared by all frames in the segment for detection sub-network and $ \feat{\timp}{k} $, the impression feature containing long-term temporal information. For the first segment $ \vseg{0} $, $ \feat{\timp}{0} $ and $ \feat{\ttsk}{0} $ are identical to $ \feat{}{0} $. For $ \vseg{1} $, $ \feat{\ttsk}{1} $ is a weighted combination of $ \feat{\timp}{0} $ and $ \feat{}{1} $. The aggregation unit $ \agg $ uses a tiny FCN $ \nq $ to generate position-wise weight maps. Generally, larger weights are assigned to the feature with better quality. This is concluded as

\begin{align}
\feat{\timp}{0}, \feat{\ttsk}{0} &= \feat{}{0}, \\
1-w_{1}, w_{1} &= \text{softmax}({\nq({\feat{\timp}{0}}'), \nq(\feat{}{1})}), \\
\feat{\ttsk}{1} &= (1-w_{1})\cdot {\feat{\timp}{0}}' + w_{1} \cdot \feat{}{1}. \label{eq.aggregate}
\end{align}

Notice that such quality is not a handcrafted metric, instead it's learned by end-to-end training to minimize task loss. We observe that when $ \vframe{\tkey}{k} $ is deteriorated by motion blur or defocus, $ \feat{}{k} $ gets lower quality score, as shown in Figure~\ref{fig.score_demo}. Also notice that the aggregation of cross-segment features is not simply adding them up. Former practice~\cite{zhu2017flow} shows that due to spatial misalignment in video frames, naive weighted mean yields worse results. Here we use flow-guided aggregation. Specifically, we first calculate the flow field of $ \vframe{\tkey}{0} $ and $ \vframe{\tkey}{1} $, then perform spatial warping accordingly on $ \feat{\timp}{0} $ to align it with $ \vframe{\tkey}{1} $, getting $ {\feat{\timp}{0}}' $; the fusion is then done with $ {\feat{\timp}{0}}' $ and $ \feat{}{1} $ to generate $ \feat{\ttsk}{1} $. $ \feat{}{1} $ and $ \feat{\ttsk}{1} $ are then mingled to get $ \feat{\timp}{1} $:

\begin{equation}
\feat{\timp}{1} = (1-\mmgt) \cdot \feat{}{1} + \mmgt \cdot \feat{\ttsk}{1}. \label{eq.mem_gate}
\end{equation}

Here a constant factor $ \mmgt $ controls the contribution of $ \feat{\ttsk}{1} $. $ \mmgt $ serves as a gate to control the memory of the framework (detailed in Figure~\ref{fig.mem_weights}). If set $ \mmgt $ to 0, $ \feat{\timp}{k} $ will only contain information of $\vframe{key}{k-1}$. The procedure keeps going on until all frames in a video get processed. 

By iteratively absorbing every keyframe feature, the impression feature contains visual information in a large time span. The weighted aggregation of $ \feat{}{k} $ and $ \feat{\timp}{k-1} $ can be seen as a balancing between memory and new information, depending on the quality of the new incoming keyframe. When the new keyframe gets deteriorated, the impression feature compensate for the subsequent weak feature, helping infer bounding box and class information through low-level visual clue such as color distribution. On the other hand, the impression feature also keeps getting updated. Since sharp and clear frames get higher scores, they contribute more to an effective impression. Compared to exhaustively aggregating all nearby features in a fixed range for every frame, our framework is more natural and elegant. The whole process is summarized in Algorithm~\ref{alg.inference}.

\begin{figure}
	\begin{center}
		\includegraphics[width=0.9\linewidth]{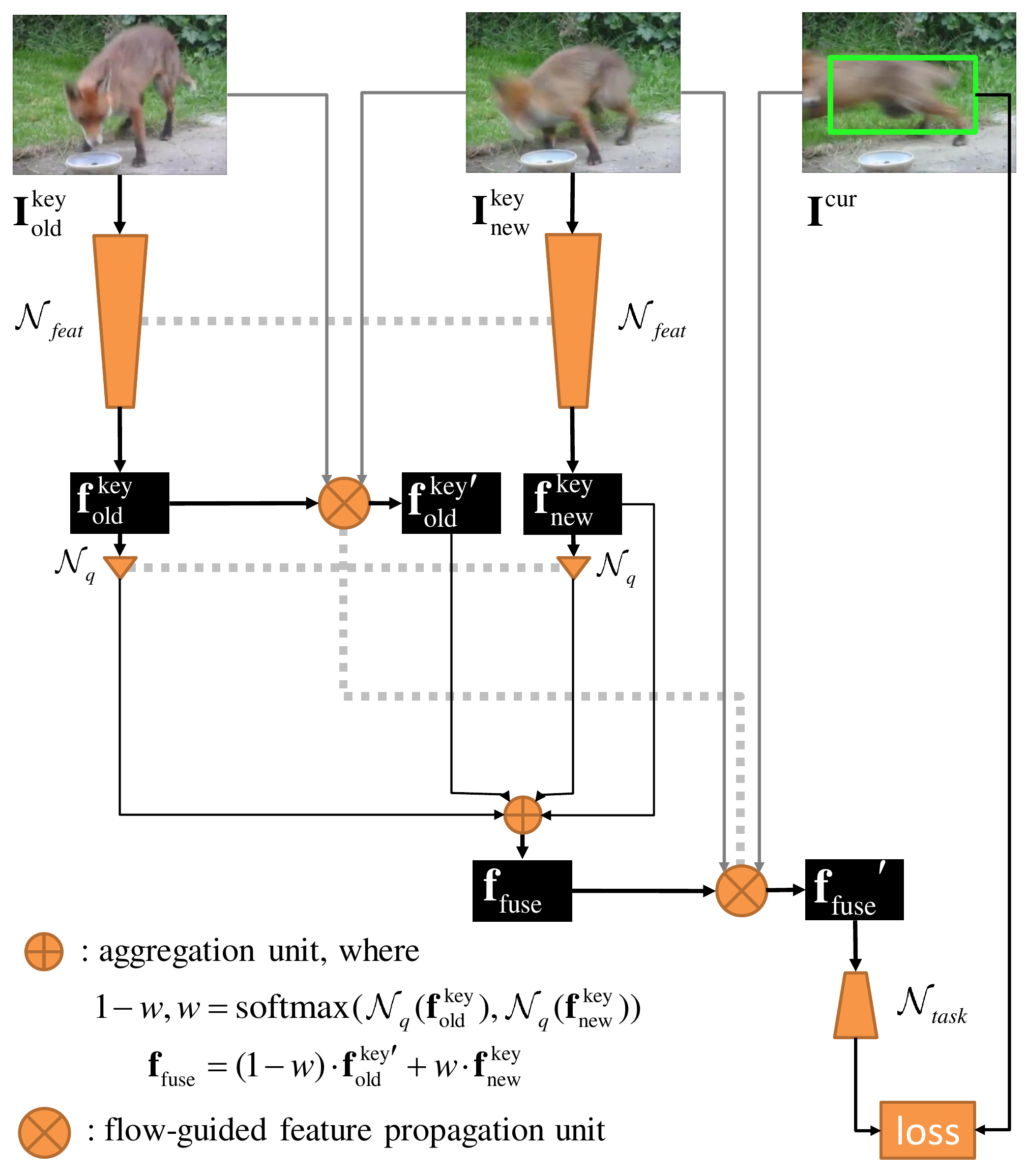}
	\end{center}
	\caption{Training framework of Impression Network. The data flow is marked by solid lines. Components linked with dashed lines share weights. The working condition of inference stage is simulated with video frame triplets. All components are optimized end-to-end. }
	\label{fig.impression_train}
\end{figure}

\subsection{Impression Network Training}

The training procedure of Impression Network is rather simple. With video data provided, a standard single-image object detection pipeline can be transfered to video tasks with slight modifications. The end-to-end training framework is illustrated in Figure~\ref{fig.impression_train}. 

During training, each data batch contains three images $ {\vframe{}{k+d_0}, \vframe{}{k}, \vframe{}{k+d_1}} $ from a same video sequence. $ d_0 $ and $ d_1 $ are random offsets whose ranges are controlled by segment length $ \kfint $. Typically, $ d_0 $ lies in $ [-\kfint, -0.5\kfint] $, while $ d_1 $ falls into $ [-0.5\kfint, 0.5\kfint] $. This setting is coherent with the inference phase, as $ \vframe{}{k} $ represents an arbitrary frame from segment $ \vseg{n} $, $ \vframe{}{k+d_1} $ for keyframe of current segment $ \vframe{\tkey}{n} $, while $ \vframe{}{k+d_0} $ stands for the previous keyframe. For simplicity, the three images are dubbed as $ \{ \vframe{\tkey}{\told}, \vframe{\tcur}{}, \vframe{\tkey}{\tnew} \} $. The ground-truth at $ \vframe{\tcur}{} $ is provided as label. 

In each iteration, first, $ \nfeat $ is applied on $ \{ \vframe{\tkey}{\told}, \vframe{\tkey}{\tnew} \} $ to get their deep features $ \{ \feat{\tkey}{\told}, \feat{\tkey}{\tnew} \} $. Then, image pairs $ \{ \vframe{\tkey}{\tnew}, \vframe{\tkey}{\told} \} $ and $ \{ \vframe{\tcur}{}, \vframe{\tkey}{\tnew} \} $ are fed into the flow network, yielding optical flow fields $ \ff{\told \to \tnew} $ and $ \ff{\tnew \to \tcur} $, respectively. Flow-guided warping unit then use $ \ff{\told \to \tnew} $ to propagate $ \feat{\tkey}{\told} $ to align with $ \feat{\tkey}{\tnew} $. We denote the warped old keyframe feature as $ {\feat{\tkey}{\told}}' $. The aggregation unit weights and fuses $ \{ {\feat{\tkey}{\told}}', \feat{\tkey}{\tnew} \} $, generating $ \feat{}{\text{fuse}} $. $ \feat{\tkey}{\told} $ in training corresponds to the impression feature in inference. This is an approximation since it only contains information of one previous keyframe. Finally, $ \feat{}{\text{fuse}} $ is warped to $ \vframe{\tcur}{} $ according to $ \ff{\tnew\to\tcur} $ to get $ {\feat{}{\text{fuse}}}' $, the task feature for a standard detection sub-network. Since all the components are differentiable, the detection loss propagates all the way back to jointly fine-tune $ \ntask $, $ \nfeat $, flow network and feature aggregation unit, optimizing task performance. Notice that single-image datasets can be fully exploited in this framework, in which case the three images are all the same. 

\subsection{Module Design}

\noindent
\textbf{Feature Network:}
We use ResNet-101 pretrained for ImageNet classification. The fully connected layers are removed. For denser feature map, feature stride is reduced from 32 to 16. Specifically, the stride of the last block is modified from 2 to 1. To maintain receptive field, A dilation of 2 is applied to convolution layers with kernel size greater than 1. A 1024-channel $ 3 \times 3 $ convolution layer (randomly initialized) is appended to reduce feature dimension. 

~\\
\noindent
\textbf{Flow-Guided Feature Propagation:}
Before aggregation, we spatially align frame features by flow-guided warping. Optical flow field is calculated first to obtain pixel-level motion path, then reference feature is warped to target frame with bilinear sampling. The procedure is defined as
\[ {\feat{\tref}{}}' = \warpp{\feat{\tref}{}}{\flowp{\vframe{\tcur}{}}{\vframe{\tref}{}}} \cdot \scl \]
where $ \vframe{\tcur}{} $ and $ \vframe{\tref}{} $ denotes target frame and reference frame respectively, $ \feat{\tref}{} $ is the deep feature of reference frame, $ {\feat{\tref}{}}' $ denotes reference feature warped to target frame, $\flow$ stands for flow estimation function, W denotes the bilinear sampler, and $\scl$ is a predicted position-wise scale map to refine warped feature. We adopt the state-of-the-art CNN-based FlowNet~\cite{dosovitskiy2015flownet,ilg2016flownet} for optical flow computation. Specifically, we use FlowNet-S~\cite{dosovitskiy2015flownet}. The flow network is pretrained on FlyingChairs dataset. The scale map has equal channel dimension with task features, and is predicted with flow field in parallel through an additional $ 1 \times 1 $ convolution layer attached to the top of FlowNet-S. The new layer is initialized with weights of all zeros and fixed biases of all ones. The implementation of bilinear sampling unit has been well described in~\cite{jaderberg2015spatial,dai2017deformable,zhu2016deep}. It is fully differentiable.

~\\
\noindent
\textbf{Aggregation Unit:}
The aggregation weights of features are generated by a quality estimation network $\nq$. It has three randomly initialized layers: a $ 3\times3\times256 $ convolution, a $1\times1\times16$ convolution and a $1\times1\times1$ convolution. The output is a position-wise raw score map which will be applied on each channel of task feature. Raw score maps of different features are normalized by softmax function to sum up to one. We then multiply the score maps with features and sum them up to obtain the fused feature as Eq.~\ref{eq.aggregate}.

~\\
\noindent
\textbf{Detection Network:}
We use the state-of-the-art R-FCN as detection sub-network. RPN and R-FCN are attached to the 1024-channel convolution of the feature network, using the first and second 512 channels respectively. RPN uses 9 anchors and generates 300 proposals for each image. We use $ 7\times7 $ groups position-sensitive score maps for R-FCN.

\subsection{Runtime Complexity Analysis}

The ratio of inference time of our method to that of per-frame evaluation is:
\[ r = \frac{\cmpx{\agg} + \kfint \times (\cmpx{\warp}+\cmpx{\flow}+\cmpx{\ntask}) + \cmpx{\nfeat}}
            { \kfint \times (\cmpx{\nfeat}+\cmpx{\ntask}) } \]
In each segment of length $\kfint$ , Impression Network requires: 1. $\kfint$ flow warping $(\cmpx{\warp}+\cmpx{\flow})$ in total, one for impression feature propagation and $\kfint-1$ for non-key frame detection; 2. One feature fusion operation $(\cmpx{\agg})$; 3. One feature network inference for keyframe feature; 4. $\kfint$ detection subnetwork inference. In comparison, per-frame solution takes $\kfint$ $ \nfeat $ and $\kfint$ $\ntask$ inference. Notice that compared to $\nfeat$ (Resnet-101 in our practice) and FlowNet, the complexity of $\agg$, $\warp$ and $\ntask$ are negligible. So the ratio can be approximated as:
\[ r \approx 
     \frac{\cmpx{\flow}}
          {\cmpx{\nfeat}} + 
     \frac{1}
          {\kfint} \]
In practice, the flow network is times smaller than Resnet-101, while $\kfint$ is large ($\geq10$) to reduce redundancy. This suggests that unlike existing feature aggregation method like FGFA, Impression Network can perform multi-frame feature fusion while maintaining a noticeable speedup over per-frame solution. 


\begin{figure}
	\begin{center}
		\includegraphics[width=0.9\linewidth]{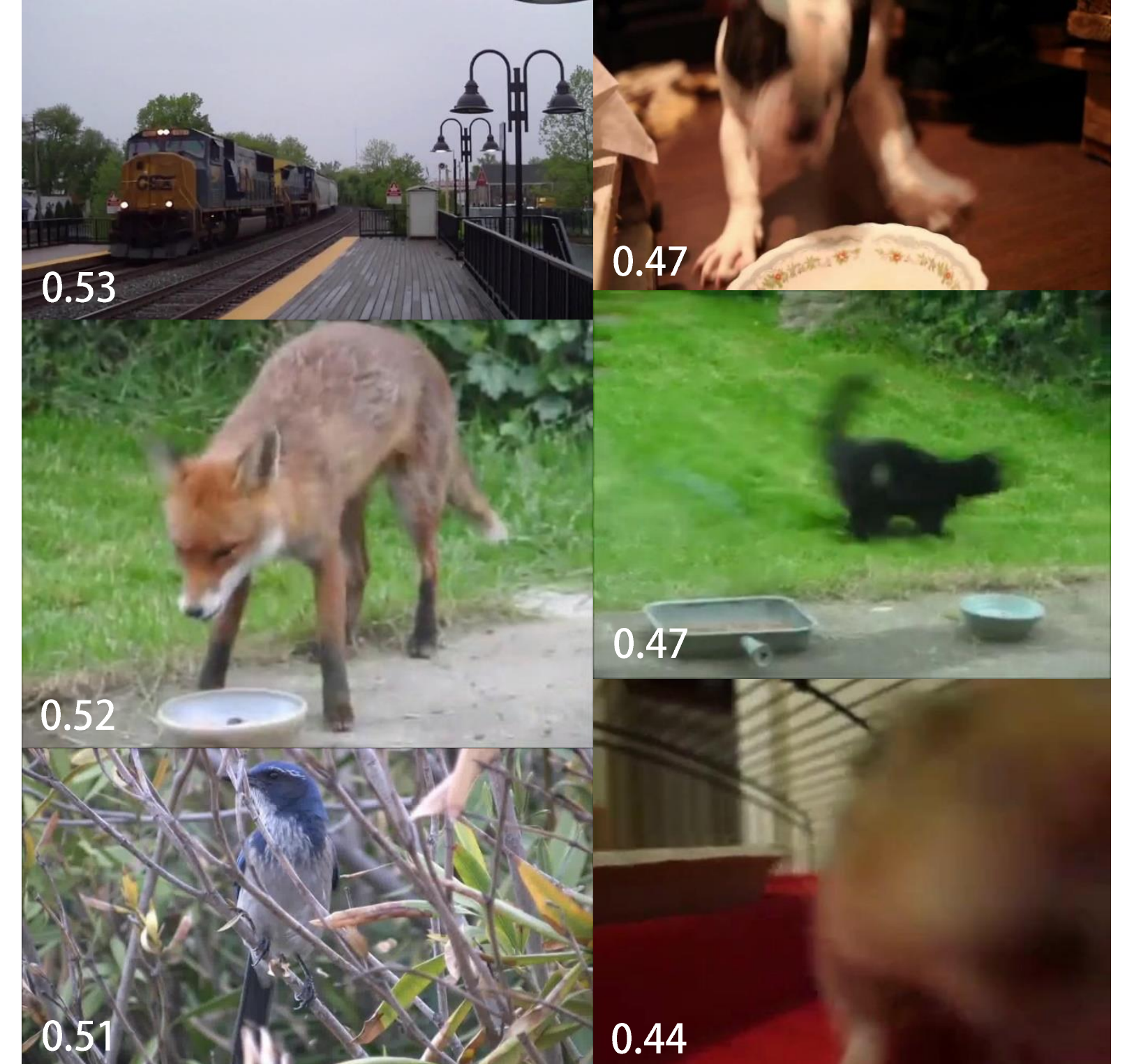}
	\end{center}
	\caption{Examples of frames assigned with different aggregation weights. The white number is the spatially averaged pixel-wise weight $w_k$ in algorithm~\ref{alg.inference}. Consistent with intuition, the scoring FCN $\nq$ assigns larger weights to sharp and clear frames. }
	\label{fig.score_demo}
\end{figure}

\begin{figure*}[t]
	\begin{center}
		\includegraphics[width=1.0\linewidth]{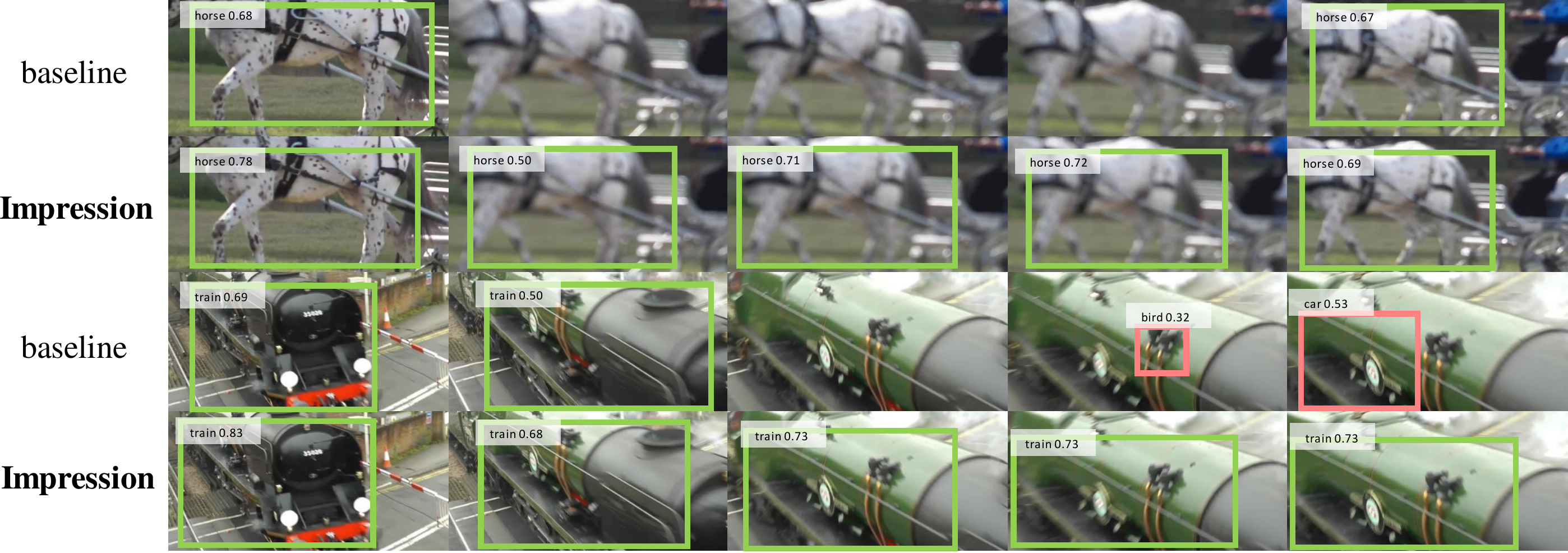}
	\end{center}
	\caption{Examples where Impression Network outperforms per-frame baseline (standard ResNet-101 R-FCN). Green boxes are true positives while red ones are false positives.}
	\label{fig.comp_baseline}
\end{figure*}

\section{Experiments}

\subsection{Experiment Setup}
\noindent
\textbf{ImageNet VID dataset~\cite{russakovsky2015imagenet}:}
It is a large-scale video object detection dataset. There are 3862, 555 and 937 snippets with frame rates of 25 and 30 in training, validation and test sets, respectively. All the video snippets are fully-annotated. Imagenet VID dataset has 30 object categories, which is a subset of the Imagenet DET dataset. In our experiments, following the practice in~\cite{kang2017t,lee2016multi,zhu2016deep,zhu2017flow}, model are trained on the training set, while evaluations are done on the validation set with the standard mean average precision (mAP) metric.

~\\
\noindent
\textbf{Implementation Details:}
Our training set consists of the full ImageNet VID train set, together with images from ImageNet DET train set. Only the same 30 categories are used. As mentioned before, each training batch contains three images. If sampled from DET, all images are same. In both training and testing, images are resized to have the shorter side of 600 and 300 pixels for the feature network and the flow network, respectively. The whole framework is trained end to end with SGD, where 120K iterations are performed on 8 GPUs. The learning rate is $ 10^{-3} $ for the first 70K iterations, then reduced to $ 10^{-4} $ for the remaining 50K iterations. For clear comparison, no bells-and-whistles like multi-scale training and box-level post-processing are used. Inference time is measured on a Nvidia GTX 1060 GPU. 

\setlength{\tabcolsep}{8pt}
\renewcommand{\arraystretch}{1.2}
\begin{table}
	\centering
	\small
	\begin{tabular}{c|c|c|c|c|c}
		\hline
		methods & (a) & (b) & (c) & \textbf{(d)} & (e) \\
		\hline
		sparse feature?   &	&	\checkmark	&	\checkmark	&	\checkmark	&	\checkmark	\\
		impression?  &  &  &	\checkmark	&	\checkmark	&	\checkmark	\\
		quality-aware?  &  &	&  & \checkmark	& \checkmark \\
		end-to-end?   & \checkmark	& \checkmark & \checkmark &	\checkmark &  \\
		\hline
		mAP (\%)  & $74.2$ & $73.6$ & $75.2$ &  $\mathbf{75.5}$ & $70.3$\\
		\hline
		runtime (ms) & 156 & 48 & 50 & $\mathbf{50}$ & 50 \\
		\hline 
	\end{tabular}
	\caption{ Accuracy and runtime of different approaches. }
	\label{tab.ablation_main}
\end{table}

\subsection{Ablation Study}

\noindent
\textbf{Architecture Design:}
Table~\ref{tab.ablation_main} summarizes main experiment results. It shows a comparison of single-frame baseline, Impression Network and its variants. 

\emph{Method (a)} is the standard ResNet-101 R-FCN applied frame by frame to videos. The accuracy is close to the $ 73.9\% $ mAP reported in ~\cite{zhu2016deep}, which shows its validity as a strong baseline for our evaluations. The runtime is a little bit faster, probably due to differences in implementation environment. The $\approx$6fps inference speed is insufficient for real-time applications, where typically a speed of $\geq$15fps is required.

\emph{Method (b)} is a variant of \emph{Method (a)} with sparse feature extraction. In this approach, videos are divided into segments of $ \kfint $ frames. Only one keyframe in each segment will be passed through the feature network for feature extraction. That feature is then propagated to other frames with optical flow. Finally, the detection sub-network generates results for every frame. The structure is identical to a Deep Feature Flow framework for video object detection ~\cite{zhu2016deep}. Specifically, $ \kfint $ is set to 10 for all experiments in this table. We select the 5th frame as keyframe, because this minimizes average feature propagation distance, thus reduces the error introduced and improves accuracy (explained later). Compared to per-frame evaluation, there's a minor accuracy drop of $ 0.6\% $, mainly because of lessened information, as well as errors in flow-guided feature warping. However, the inference speed remarkably increases to 21fps, proving that sparse feature extraction is an efficient way to trade accuracy for speed.

\emph{Method (c)} is a degenerated version of Impression Network. Keyframe features are iteratively fused to generate the impression feature, but without quality-aware weighting unit. The weights in Eq.~\ref{eq.aggregate} are naively fixed to $ 0.5 $. For all experiments here, the memory gate $ \mmgt $ in Eq.~\ref{eq.mem_gate} is set to $ 1.0 $. With information of previous frames fused into current task feature, mAP increases for $1.0\%$ over per-frame baseline. Notice that sparse feature extraction is still enabled here, which proves that 1.the computational redundancy of per-frame evaluation is huge; 2.such redundancy is not necessary for higher accuracy. Due to the one additional flow estimation for each segment, the framework slows down a little bit, yet still runs at a real-time-level 20fps.

\emph{Method (d)} is the proposed Impression Network. Here the aggregation unit uses the tiny FCN to generate position-wise weights. Through end-to-end training, the sub-network learns to assign smaller weights to features of deteriorated frames, as shown in Figure~\ref{fig.score_demo}. Experiment on ImageNet VID validation set shows the $w_k$ in algorithm~\ref{alg.inference} obeys a normal distribution of $\mathcal{N}(0.5, 0.016^2)$. Quality-aware weighting brings another $0.3\%$ mAP improvement, mainly because of the increment of valid information. Overall, Impression Network increases mAP by $1.3\%$ to $75.5\%$, comparable to exhaustive feature aggregation method~\cite{zhu2017flow}, while significantly faster, running at 20fps. Impression Network shows that, if redundancy and temporal information are properly handled, the speed and accuracy of video object detection can actually be simultaneously improved. Examples are shown in Figure~\ref{fig.comp_baseline}.

\emph{Method (e)} is Impression Network without end-to-end training. The feature network is trained in single-image detection framework, same to that in \emph{Method (a)}. The flow network is the off-the-shelf FlyingChairs pretrained FlowNet-S~\cite{dosovitskiy2015flownet}. Only the weighting and detection sub-networks learn during training. This clearly worsen the mAP, showing the importance of end-to-end optimization. 

\begin{figure}
	\begin{center}
		\includegraphics[width=0.9\linewidth]{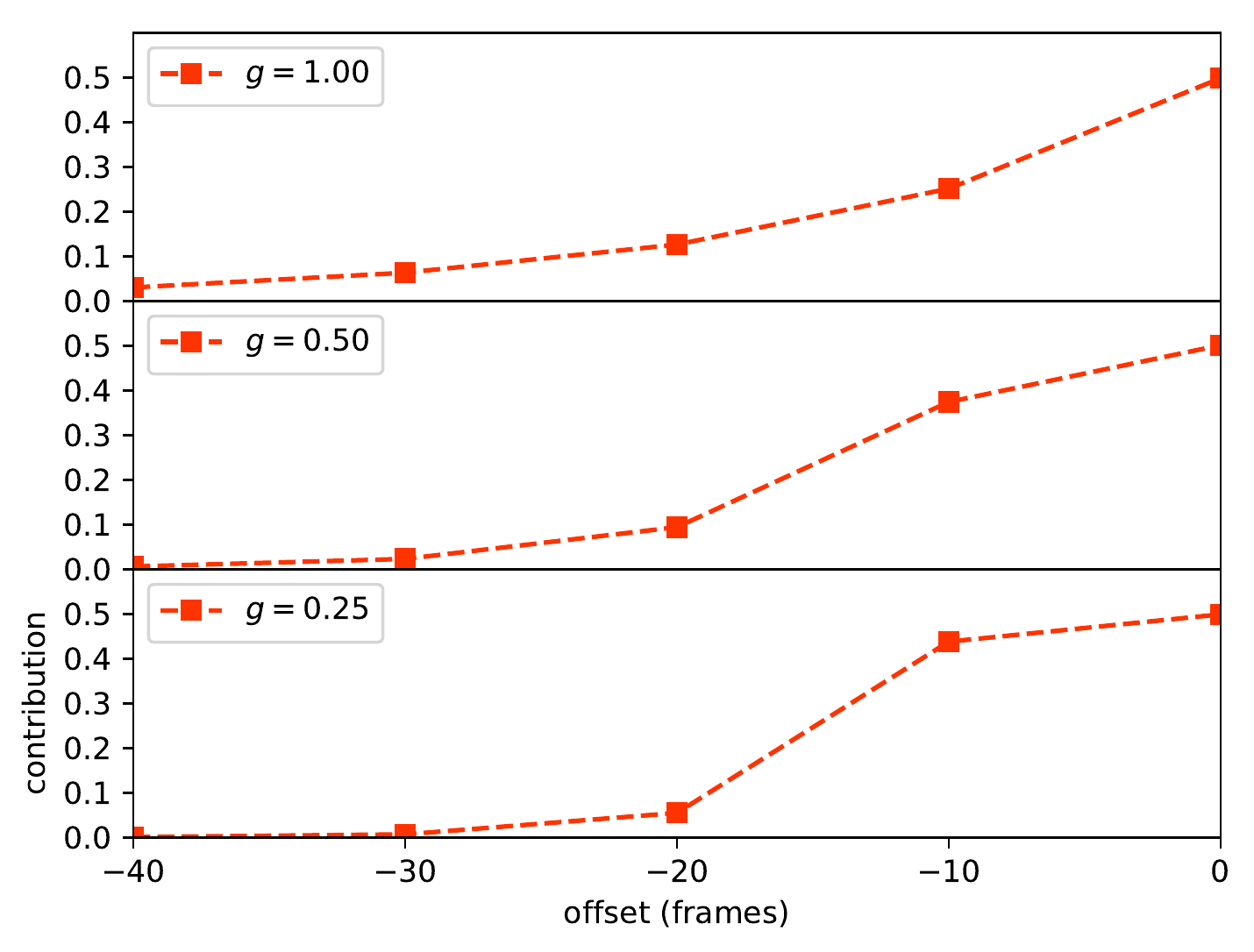}
	\end{center}
	\caption{Averaged contribution of previous keyframes to current detection at different memory gate $\mmgt$. When $\mmgt$ is $1.0$, the contribution smoothly decreases as offset grows. As $\mmgt$ decreases, the impression gets increasingly occupied by the nearest keyframe, while the contribution of earlier ones rapidly shrinks to $0$.}
	\label{fig.mem_weights}
\end{figure}

\begin{figure}
	\begin{center}
		\includegraphics[width=0.9\linewidth]{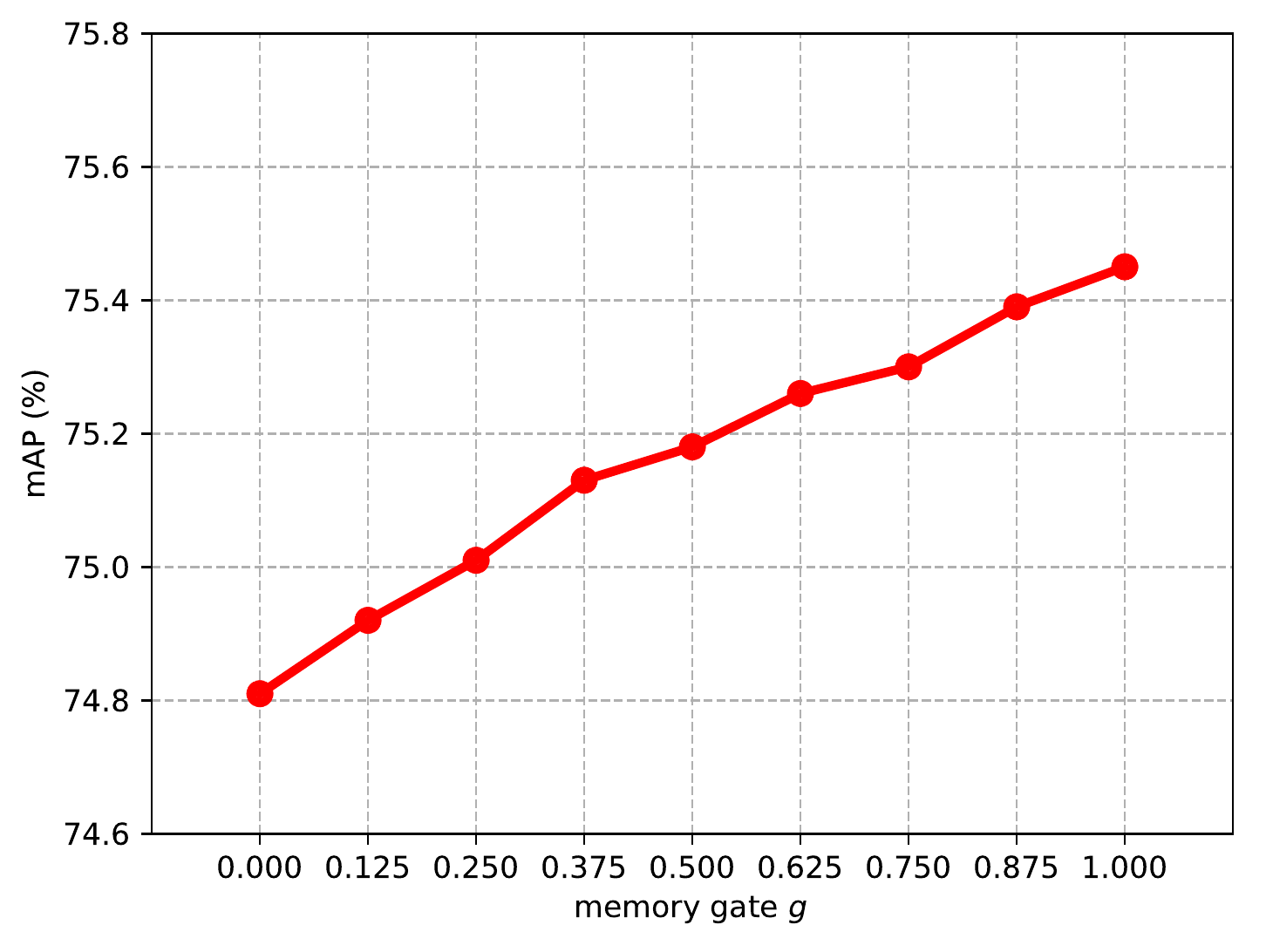}
	\end{center}
	\caption{mAP at different $\mmgt$ values. Although it's not exactly how the network is trained, enabling long-range aggregation do brings noticeable improvement.}
	\label{fig.mem_curve}
\end{figure}

~\\
\noindent
\textbf{The Influence of Memory Gate:}
As shown in Eq.~\ref{eq.mem_gate}, the memory gate $\mmgt$ controls the component of impression features. Here we study its influence on mAP. Experiment settings are the same as \emph{Method (d)} in Table~\ref{tab.ablation_main}, except that $\mmgt$ varies from $0.0$ to $1.0$. Figure~\ref{fig.mem_weights} shows the average contribution of previous keyframes to current detection at different $\mmgt$ values. It can be seen that $\mmgt$ controls the available range of temporal information. When set to $0.0$, the impression feature consists solely of the previous key feature, just like how the framework is trained; while setting $\mmgt$ to 1.0 leads in more temporal information. Figure~\ref{fig.mem_curve} shows the mAP of different $\mmgt$ setting. Apparently, larger $\mmgt$ benefits accuracy. The involvement of long-range feature aggregation may help detection in longer series of low-quality frames. 

\setlength{\tabcolsep}{8pt}
\renewcommand{\arraystretch}{1.2}
\begin{table}
	\centering
	\small
	\begin{tabular}{c|c|c}
		\hline
		keyframe id & $\bar{d}$ (frames) & mAP (\%) \\
		\hline \hline
		0  & 5.5 & 73.9 \\
		\hline
		1  & 4.7 & 74.4\\
		\hline
		2  & 4.1 & 74.9\\
		\hline
		3  & 3.7 & 75.2\\
		\hline
		4  & 3.5 & 75.5\\
		\hline
		5  & 3.5 & 75.5\\
		\hline
	\end{tabular}
	\caption{Average propagation distance and mAP at different keyframe selections. Other settings are same as \emph{Method (d)} in Table~\ref{tab.ablation_main}. Because of the symmetry, only id 0-5 is shown.}
	\label{tab.select_key}
\end{table}

~\\
\noindent
\textbf{Different Keyframe Selection:}
In aforementioned experiments, to reduce error, we select the central frame of each segment as keyframe. Here we explain this and compare different keyframe scheduling. Flow-guided feature warping introduces error, and as shown in ~\cite{zhu2016deep}, the error has positive correlation with propagation distance. This is because that larger displacement increases the difficulty of pixel-level matching. Hence, we take average feature propagation distance $ \bar{d} $ as a metric for flow error, and seek the way to minimize it. $ \bar{d} $ is calculated as:  
\[
\bar{d}=
\begin{dcases}
\frac{\sum_{d=1}^{\kfint-1} d + \kfint}{\kfint}, & k=0, \kfint-1 \\
\frac{\sum_{d=1}^{k} d+\sum_{d=1}^{\kfint-1-k} d + \kfint}{\kfint}, & 0<k<\kfint-1
\end{dcases}
\]
where $d$ is propagation distance, $k$ is the id of keyframe, and $\kfint$ is segment length. Key feature needs to be propagated to non-key frames, and there's also an impression feature propagation of distance $\kfint$. Apparently there's an optimal $k$ to minimize $\bar{d}$: 
\begin{equation}
\mathop{\arg\min}_{k}(\bar{d}) = \frac{\kfint-1}{2}
\end{equation}
which shows that the central frame is the best. Table~\ref{tab.select_key} shows mAPs at different keyframe selections, coherent with our assumption. Notice that selecting the first frame enables strict real-time inference, while selecting the central frame brings a slight latency of $\kfint/2$ frames. This can be traded-off according to application needs. 

\begin{figure}
	\begin{center}
		\includegraphics[width=0.9\linewidth]{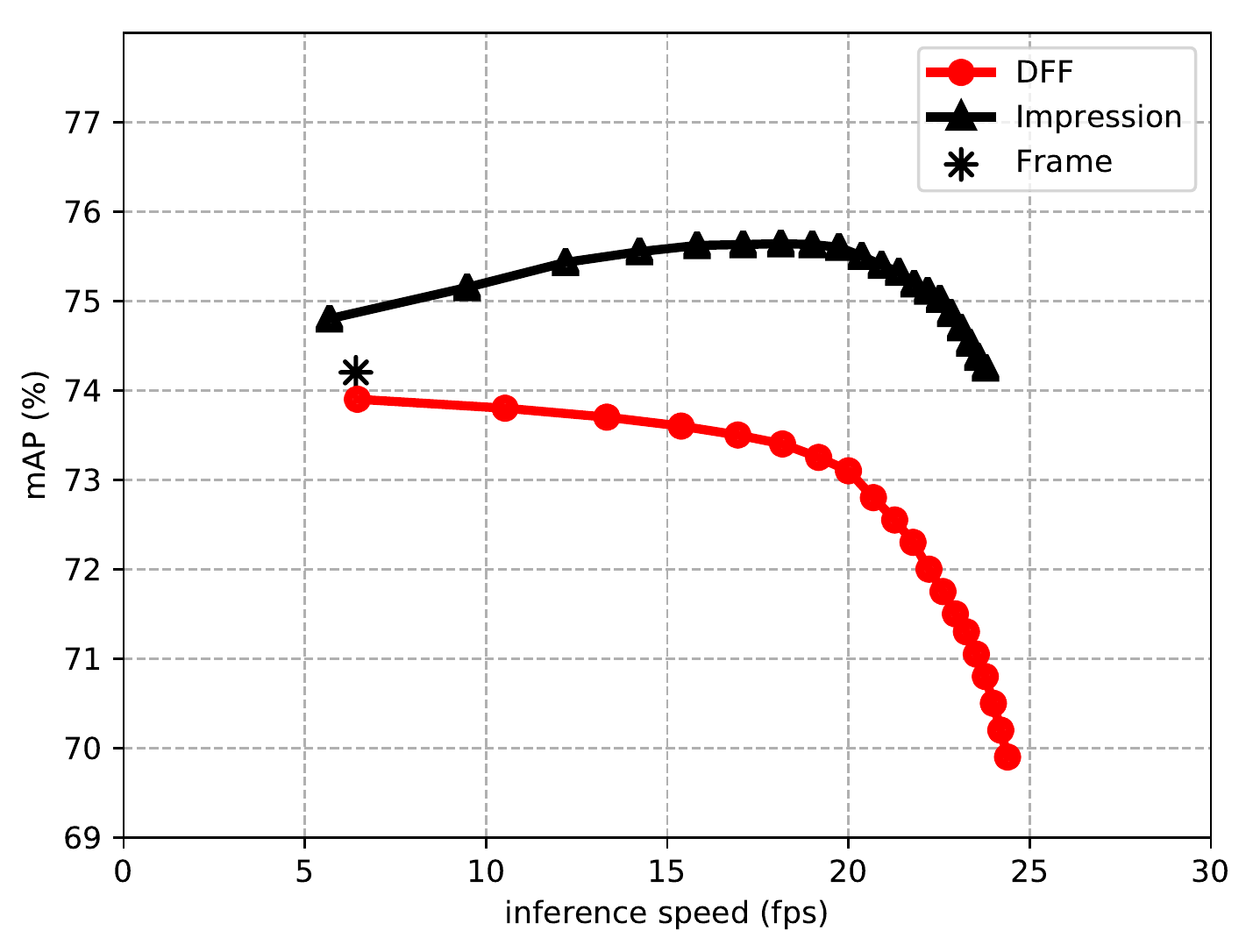}
	\end{center}
	\caption{Comparing speed-accuracy curves of Deep Feature Flow (DFF) and Impression Network (Impression). Both using ResNet-101 as feature network and FlowNet-S as flow network.}
	\label{fig.comp_dff}
\end{figure}

\setlength{\tabcolsep}{8pt}
\renewcommand{\arraystretch}{1.2}
\begin{table}
	\centering
	\small
	\begin{tabular}{c|c|c}
		\hline
		method & mAP (\%) & runtime (ms)\\
		\hline \hline
		FGFA  & 76.3 & 733 \\
		\hline
		FGFA-fast  & 75.3 & 356\\
		\hline
		Impression Network  & 75.5 & 50\\
		\hline
	\end{tabular}
	\caption{Comparison with aggregation-based method FGFA and its faster variant.  Settings are same as \emph{Method (d)} in Table~\ref{tab.ablation_main}.}
	\label{tab.comp_fgfa}
\end{table}

\subsection{Compare with Other Feature-Level Methods}
We compare Impression Network with other feature-level video object detection methods. In Figure~\ref{fig.comp_dff}, we compare the speed-accuracy curve of Impression Network and Deep Feature Flow~\cite{zhu2016deep}. Per-frame baseline is also marked. Segment length $\kfint$ varies from 1 to 20. Apparently, Impression Network is more accurate than per-frame solution even in high-speed zone. Similar to Deep Feature Flow, it also offers a smooth accuracy-speed trade-off as $\kfint$ varies. The accuracy drops a little when $\kfint$ gets close to 1, which is reasonable because Impression Network is trained for aggregating sparse frame features. Dense sampling limits aggregation range and result in a less useful impression.

Table~\ref{tab.comp_fgfa} compares Impression Network with Flow-Guided Feature Aggregation and its faster variant. Both are described in~\cite{zhu2017flow}. FGFA is the standard version with a fusion radius of 10, and FGFA-fast is the accelerated version. It only calculates flow fields for adjacent frames, and composite them for non-adjacent pairs. This comparison shows that the accuracy of Impression Network is on par with the best aggregation-based method, yet being much more efficient.

\section{Conclusion and Future Work}
This work presents a fast and accurate feature-level method for video object detection. The proposed Impression mechanism explores a novel scheme for feature aggregation in videos. Since Impression Network works at feature stage, it's complementary to existing box-level post-processing methods like Seq-NMS~\cite{han2016seq}. For now we use FlowNet-S~\cite{dosovitskiy2015flownet} to guide feature propagation for clear comparison, while more efficient flow algorithms~\cite{ilg2016flownet} exist and can surely benefit our method. We use fixed segment length for simplicity, while a adaptively varying length may schedule computation more reasonably. Moreover, as a feature-level method, Impression Network inherits the task-independence, and has the potential to tackle image degeneration problem in other video tasks. 

{\small
\bibliographystyle{ieee}
\bibliography{egbib}
}

\end{document}